\documentclass[sigconf]{acmart}

\setcopyright{acmlicensed}
\copyrightyear{2025}
\acmYear{2025}
\acmConference[HOTMOBILE '25]{The 26th International Workshop on Mobile Computing Systems and Applications}{February 26--27, 2025}{La Quinta, CA, USA}
\acmBooktitle{The 26th International Workshop on Mobile Computing Systems and Applications (HOTMOBILE '25), February 26--27, 2025, La Quinta, CA, USA}
\acmDOI{10.1145/3708468.3711890}
\acmISBN{979-8-4007-1403-0/25/02}

\usepackage{multirow}
\usepackage{algorithm}
\usepackage{algpseudocode}
\usepackage{subcaption}
\usepackage{makecell}

\begin{document}

\title{\textsc{GreenAuto}: An Automated Platform for Sustainable AI Model Design on Edge Devices}

\author{Xiaolong Tu}
\email{xtu1@gsu.edu}
\authornotemark[1]
\affiliation{%
  \institution{Georgia State University}
  \city{Atlanta}
  \state{Georgia}
  \country{USA}
}

\author{Dawei Chen}
\email{dawei.chen1@toyota.com}
\authornotemark[2]
\affiliation{%
  \institution{Toyota InfoTech Labs}
  \city{Mountain View}
  \state{California}
  \country{USA}
}

\author{Kyungtae Han}
\email{kt.han@toyota.com}
\authornotemark[2]
\affiliation{%
  \institution{Toyota InfoTech Labs}
  \city{Mountain View}
  \state{California}
  \country{USA}
}

\author{Onur Altintas}
\email{onur.altintas@toyota.com}
\authornotemark[2]
\affiliation{%
  \institution{Toyota InfoTech Labs}
  \city{Mountain View}
  \state{California}
  \country{USA}
}

\author{Haoxin Wang}
\email{haoxinwang@gsu.edu}
\authornotemark[1]
\affiliation{%
  \institution{Georgia State University}
  \city{Atlanta}
  \state{Georgia}
  \country{USA}
}



\renewcommand{\shortauthors}{Xiaolong Tu et al.}

\begin{abstract}

We present \textsc{GreenAuto}, an end-to-end automated platform designed for sustainable AI model exploration, generation, deployment, and evaluation. \textsc{GreenAuto} employs a Pareto front-based search method within an expanded neural architecture search (NAS) space, guided by gradient descent to optimize model exploration. Pre-trained kernel-level energy predictors estimate energy consumption across all models, providing a global view that directs the search toward more sustainable solutions. By automating performance measurements and iteratively refining the search process, \textsc{GreenAuto} demonstrates the efficient identification of sustainable AI models without the need for human intervention.
\end{abstract}


\maketitle

\section{Introduction}


Sustainable Artificial Intelligence (AI) refers to the development and application of AI technologies aimed at addressing environmental challenges and promoting sustainability \cite{wu2022sustainable, tu2023deepen2023}. This field spans two key areas: the sustainability of AI itself and the use of AI for advancing sustainability \cite{van2021sustainable, vinuesa2020role}. While there is growing interest in utilizing AI to support environmental sustainability, research focusing on AI’s own sustainability remains relatively scarce.
On-device learning, especially on mobile and edge devices, is gaining traction for model personalization and improved data privacy. However, the sustainability of on-device learning often receives little attention. One reason is that it is not intuitively measurable and cannot be assessed using conventional metrics. Another challenge lies in the typically short runtime of models on edge devices, which makes tracking sustainability difficult. Since most edge devices are electrically powered, and electricity generation is a significant source of carbon dioxide emissions \cite{electricitygeneration}, energy consumption serves as a practical and quantifiable proxy for evaluating sustainability. The importance of energy efficiency becomes even more apparent when considering the scale of edge devices. By 2030, the number of edge devices is projected to reach 40 billion \cite{estimate}, their collective carbon footprint could surpass 292.42 million metric tons annually, 1.145 times France's carbon emissions in 2023.
A major contributor to this carbon footprint is the runtime of AI models on these edge devices, such as video analytics and voice assistants \cite{wang2020energy, wang2022leaf+}. As AI-driven applications become more prevalent, their cumulative energy consumption across billions of devices will have a significant impact on global carbon emissions. 
Addressing this pressing issue requires a deeper understanding of how AI models consume energy during runtime and how to optimize their energy efficiency for specific edge hardware. This optimization is crucial for reducing the environmental impact of AI at scale. As the prevalence of on-device learning grows, integrating sustainability into its design and evaluation processes is essential to align technological advancements with environmental goals.

 To address this, we design \textsc{GreenAuto}, an end-to-end automated platform to optimize AI models' sustainability from model itself and the model search process. We use energy consumption and converted carbon emissions as representatives, offering a straightforward approach for measurement and evaluation of the sustainability. \textsc{GreenAuto} is designed based on energy predictors, integrates with NAS search space, Pareto front-based model search, and automated energy measurements in a completed pipeline to search the sustainability models automatically. It demonstrates strong potential for sustainable AI model design by streamlining the search process and generating energy-efficient models. \textsc{GreenAuto} offers several unique features:

\textit{Sustainable model search:} \textsc{GreenAuto} extends existing NAS spaces and uses pre-trained energy predictors to estimate energy consumption for all model candidates. Pareto front optimization and gradient descent techniques are employed to efficiently balance energy efficiency and accuracy, significantly reducing training time and human intervention.

\textit{Automated energy measurement:}
\textsc{GreenAuto} automates energy measurement on edge devices without human intervention. It iteratively and interactively collects energy and latency data to inform the model search. 

\textit{Scalability and flexibility:} \textsc{GreenAuto} is modular and reconfigurable, allowing researchers to replace components, adjust parameters, and prioritize different objectives (e.g., accuracy or energy efficiency). This flexibility enables easy customization for diverse research needs and use cases.
\section{System Design}

\begin{figure*}[t]
  \centering
  {\includegraphics[width=0.9\linewidth]{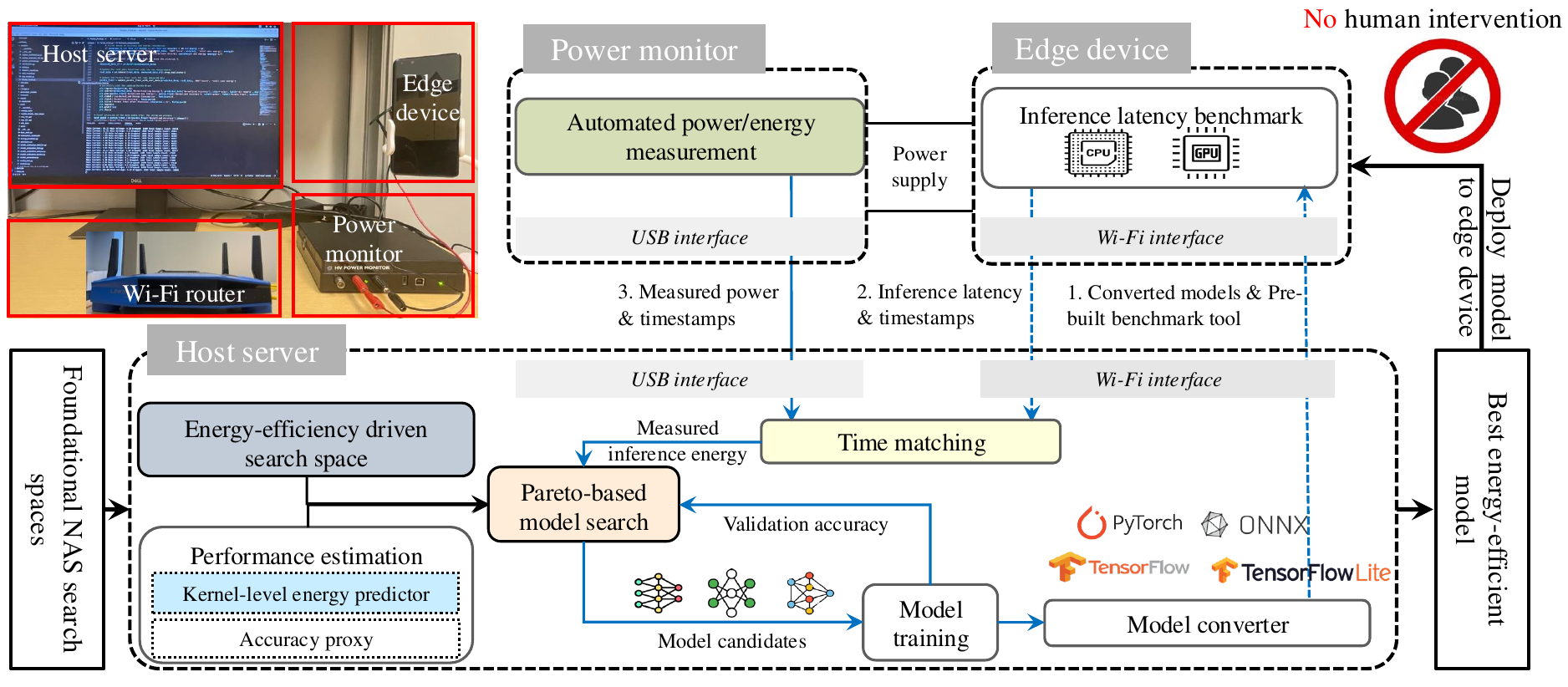}}
  \caption{System overview of \textsc{GreenAuto}, operating autonomously without human intervention.} 
  \Description{system pipeline}
  \label{fig:system pipeline}
\end{figure*}

In this section, we introduce \textsc{GreenAuto}, our end-to-end platform designed to automate the environmentally sustainable design of DNN models for edge devices.
\textsc{GreenAuto} contains three essential components: (1) \textbf{energy efficiency driven model search space} (\S \ref{sec: search space}) that includes over 900,000 DNN models tailored for energy efficiency, (2) \textbf{Pareto front-based model search algorithm} (\S \ref{ssc:Pareto}) that, combined with gradient descent-based model sampling, efficiently identifies promising model candidates from the search space, and (3) \textbf{automated energy measurement function} (\S \ref{ssc:measure}) that leverages an external power monitor to capture accurate power consumption data during model execution.

Fig. \ref{fig:system pipeline} illustrates the end-to-end pipeline of \textsc{GreenAuto} and its implementation. \textsc{GreenAuto} operates autonomously without requiring human intervention.
The process begins by generating an energy efficiency-driven model search space. We create new DNN model variants by expanding the structural parameters, such as channel size, kernel size, and stride, of foundational model architectures derived from existing neural architecture search (NAS) spaces, like NAS-Bench-201 \cite{dong2020bench}. This expanded search space focuses on models with greater potential to optimize the trade-off between accuracy and energy consumption.
Next, \textsc{GreenAuto} uses our pre-trained kernel-level energy predictors \cite{10419248} and an accuracy proxy to generate initial energy and accuracy estimates for each model. These estimates effectively accelerate the search process and reduce system overhead in identifying energy-efficient models.
We design and implement a Pareto front-based optimization algorithm, integrated with gradient descent-based model sampling in \textsc{GreenAuto}, to iteratively explore the search space with minimal overhead. By leveraging energy and accuracy estimates as a guide, the algorithm prioritizes DNN models that better balance inference accuracy and energy consumption.
After each iteration of exploration, an automated energy efficiency assessment is conducted. \textsc{GreenAuto} precisely profiles the actual energy consumption of selected model candidates when they are deployed and running on the target edge device using an external power monitor. The evaluation results from each iteration are continuously used to update the Pareto front, progressively refining the set of optimal models.
Additionally, \textsc{GreenAuto} is highly reconfigurable. By adjusting search spaces, hardware devices, or objective weights, it can easily be adapted by other researchers to meet specific optimization needs, making it a versatile tool for a wide range of applications (discussed in \S \ref{ssc:reconfigurability}).


\begin{table}[t]
\centering
\caption{Convolutional layer configurations}
\footnotesize
\resizebox{0.46\textwidth}{!}{
\begin{tabular}{l|l|l}
\toprule
Configurations   & Original space   & Expanded space \\
\hline    
Output channel                  & 16, 32, 64       & 1, 2, 3, 4, 5, 6, 7, 8, 9, 10, 16, 32, 64, 128, 256 \\\hline
Kernel size    & 1, 3   & 1, 3, 5, 7 \\\hline
Stride                  & 1       & 1, 2 \\
\bottomrule
\end{tabular}}
\label{tb:nastable}
\end{table}

\subsection{Energy Efficiency-Driven Search Space} 
\label{sec: search space}
\textsc{GreenAuto} initially expands the foundational search space to prioritize energy efficiency. For example, it extends NAS-Bench-201 \cite{dong2020bench}, a widely used cell-based search space benchmark for comparing NAS algorithms. NAS-Bench-201 offers a standardized framework for evaluating cell-based architectures.


Inspired by our previous work \cite{10419248}, we recognize that the energy consumption of a DNN is primarily influenced by both its architecture and its configuration parameters, such as input channel size ($C_{in}$), output channel size ($C_{out}$), kernel size ($KS$), and stride ($S$). 
However, NAS-Bench-201 offers limited diversity in these configurations, where for any given architecture, its $C_{in}$, $C_{out}$, and $S$ are fixed. For instance, the $C_{out}$ values are set to 16, 32, and 64 for the first, second, and third stages, respectively, and the $KS$ is restricted to 1 or 3. This lack of flexibility in configuration limits the diversity of models that can be explored, particularly with respect to optimizing for energy efficiency.
Hence, to improve the discover of energy-efficient models, \textsc{GreenAuto} expands the configurations of $C_{out}$, $KS$, and $S$ for each cell, as shown in Table \ref{tb:nastable}, creating a broader range of model variants. Our newly expanded search space contains 959,417 valid models.

DNN model exploration typically involves resource consuming tasks, such as model training and runtime performance evaluation on real devices. Energy-efficient model search is especially costly, as optimal models vary across hardware platforms. This necessitates re-deployment and re-measurement of energy consumption on each device, making the process prohibitively expensive and limiting the applicability of hardware-aware NAS. 
\textsc{GreenAuto} addresses this challenge by leveraging our pre-trained kernel-level energy predictors \cite{10419248} and a zero-cost proxy method introduced by Neural Architecture Search without Training (NASWOT Score) \cite{mellor2021neural} to generate initial energy and accuracy estimates for all model candidates within the expanded search space. Our energy predictors achieved an average prediction accuracy of 86.2\% in previous work. In this paper, with a biased search space where configurations were limited to a narrow range to prioritize energy-efficient model design. The Kendall's Tau correlation between the predicted energy and the actual measured energy of 1,179 models is 0.526, demonstrating that our predictors can provide reliable energy estimates.
These estimates collectively provide a comprehensive view of the energy-accuracy distribution across the entire search space, with minimal overhead. 

\subsection{Pareto fornt-Based Model Search}
\label{ssc:Pareto}

Building on our energy efficiency-driven search space and the corresponding energy consumption and accuracy estimates, \textsc{GreenAuto} explores DNN models that achieve better energy efficiency while maintaining or exceeding current levels of inference accuracy.
This search process can be formulated as a multi-objective optimization problem:
\begin{small}
\begin{equation}
    \begin{aligned}
      \min F(x) = (f_{1}(x), f_{2}(x), ..., f_{m}(x)): \Omega \to \mathbb{R}^m,  
    \end{aligned}
\end{equation}
\end{small}
\noindent where $x$ is an DNN model within the search space $\Omega$, and $f_{i}(x)$ denotes the $i-$th objective function to be minimized. 
The Pareto front is a widely used method for handling non-linear multi-objective problems due to its ability to identify a set of optimal solutions that balance competing objectives without requiring predefined weights. This approach is particularly advantageous in scenarios where objectives conflict, as it provides a comprehensive view of trade-offs, enabling decision-makers to select solutions that best meet their priorities.
Given two models, $x_{1}, x_{2} \in \Omega$, we say that $x_{1}$ Pareto dominates $x_{2}$ (denoted as $x_{1} \prec x_{2}$) if:
\begin{small}
\begin{equation}
    \left\{
    \begin{aligned}
    &\forall i \in \{1, \dots, m\}, f_i(x_{1}) \leq f_i(x_{2}); \\
    &\exists j \in \{1, \dots, m\}, f_j(x_{1}) < f_j(x_{2}).
    \end{aligned}
    \right.
\end{equation}
\end{small}
\noindent The first condition states that model $x_{1}$ is no worse than model $x_{2}$ across all objectives, and the second condition indicates that $x_{1}$ is strictly better than $x_{2}$ in at least one objective. $x_{1}$ is considered globally Pareto optimal on $\mathbb{R}^m$ if it is not dominated by any other model. The collection of function values $F(x)$ for all Pareto optimal models $x$ is referred to as the Pareto front $\mathcal{P}$.
In \textsc{GreenAuto}, the model's estimated energy consumption ($E_p$) and NASWOT score ($N_s$), as discussed in \S \ref{sec: search space}, serve as the two objective functions for exploring the Pareto optimal models. Prior to the search, it normalizes $E_p$ using Min-Max Normalization and $N_s$ using Log Normalization. The resulting normalized energy consumption ($E_{p_{\text{norm}}}$) and accuracy ($N_{s_{\text{norm}}}$) are expressed as:
\begin{small}
\begin{equation}
    \begin{aligned}
    E_{p_{\text{norm}}} &= \frac{E_p - \min(E_p)}{\max(E_p) - \min(E_p)}, \\
    N_{s_{\text{norm}}} &= \log(N_s).
    \end{aligned}
    \label{eq:normalization}
\end{equation}
\end{small}
During the search process, \textsc{GreenAuto} first selects $k$ candidates from the expanded search space based on the distributions of $E_p$ and $N_s$. It then trains these $k$ models to validate their accuracy and physically measures their actual energy consumption on the target edge device, establishing the initial Pareto front.
Based on the real accuracy and energy consumption of these $k$ models, \textsc{GreenAuto} calculates the initial optimal gradient direction using multiple gradient descent (MGD). The optimal gradient direction, $g^*(x)$, is given by:
$g^*(x) \propto \sum_{i=1}^{m} \lambda_i^*(x) g_i(x)$, 
where $ \left\{ \lambda_i^*(x) \right\}_{i=1}^{m} $ is the solution to:
\begin{small}
\begin{equation}
    \begin{aligned}
       g^*(x) = \min_{\{\lambda_i\}} \left\| \sum_{i=1}^{m} \lambda_i g_i(x) \right\| \quad \text{s.t.}\quad \sum_{i=1}^{m} \lambda_i = 1,\quad \lambda_i \geq 0,\quad \forall i.
    \end{aligned}
\end{equation}
\end{small}
$g_i(x)$ is the the gradient of the $i-$th objective of model $x$.

\textsc{GreenAuto} then leverages the optimal gradient direction to guide model sampling. New DNN models are selected based on the similarity between their individual gradient, $g_i(x)$, and the current optimal gradient direction, $g^*(x)$. The gradient $g_i(x)$ is derived from $E_p$ and $N_s$. The newly selected models are automatically evaluated by \textsc{GreenAuto} to update the initial Pareto front. This process is  repeated iteratively, where in each iteration, the top-$m$ models with the highest inner product values (i.e., indicating the closest alignment to the optimal gradient) are chosen to further refine the Pareto front. The iterations continue until a model that meets the desired constraints is found.
Additionally, \textsc{GreenAuto} introduces two weights, denoted as $ws_i^*(x)$, for the search process. These weights are applied to $\lambda_i^*(x)$ for each objective during exploration. By adjusting the ratios of these weights, the search process can be effectively guided toward specific scenarios, allowing certain objectives to be prioritized over others based on the desired outcome.
Our proposed model search and sampling algorithm is detailed in Algorithm \ref{algorithm:Pareto_search}. 

\begin{algorithm}[t]
\footnotesize
\caption{\small Pareto-Based Model Search and Sampling}
\label{algorithm:Pareto_search}
\begin{algorithmic}[1] 
    \State \textbf{Input:} Normalized $E_p(x)$ and $N_s(x)$
    \State \textbf{Initialize:}
    \State Select $k$ models $k \subseteq \mathcal{X}$ based on the distributions of $E_{p}(x)$ and $N_{s}(x)$
    \State Train and measure selected models $k$
    \State Generate the initial Pareto front $\mathcal{P}_0$ using actual measurements 
    \State Compute the initial optimal gradient descent direction $g^*(x)_{0} \in \mathbb{R}^2$ using the models in $k$
    \State Add model search weights $ws_i$ to $\lambda_i$ to bias the optimal gradient direction towards specific objectives
    \While {no model in $\mathcal{P}$ satisfies the predefined constraints}
        \State Select $m$ new models from $\mathcal{X} \setminus k$    based on the similarity of their gradient $g(x)$ to $g^*(x)$
        \State Perform real measurements for the selected models
        \State Update the current Pareto front $\mathcal{P}$ using the newly measured data
        \State Recompute the optimal gradient descent direction $g^*(x)$
    \EndWhile
    \State \textbf{Output:} Pareto front $\mathcal{P}^*$ that satisfies the constraints
\end{algorithmic}
\end{algorithm}

After each iteration, once the Pareto front has been updated, \textsc{GreenAuto} applies a gradient descent-based algorithm to select the best model from the Pareto front. Similarly, this process involves calculating the gradients for each point on the Pareto front and then applying two additional model selection weights, $wd_i$, to the gradients of each dimension, allowing bias toward desired model characteristics. The best model is determined by identifying the one with the smallest magnitude of weighted gradients. 
Our designed model selection algorithm is outlined in Algorithm \ref{algorithm:BestModelSelection}. 

\begin{algorithm}[t]
\footnotesize
\caption{\small Gradient Descent-Based Model Selection}
\label{algorithm:BestModelSelection}
\begin{algorithmic}[1]
    \State \textbf{Input:} Updated Pareto front $\mathcal{P}$, model selection weights $wd_i$ for each objective, where $i = 1, 2, \dots, m$
    \State \textbf{Initialize:} Set the best model candidate $x^* = \emptyset$, minimum gradient magnitude $\min\_grad = \infty$
    \For{each model $x \in \mathcal{P}$}
        \State Compute the gradient $g_i(x)$ of the model $x$ on the Pareto front
        \State Compute weighted gradient for each dimension: $ g_i^w = wd_i \cdot g_i(x)$
        \State Calculate the magnitude of the weighted gradient: $\|g_i^w \|$
        \If{$\| g_i^w \| < \min\_grad$}
            \State Update $x^* = x$
            \State Update $\min\_grad = \|g_i^w \|$
        \EndIf
    \EndFor
    \State \textbf{Output:} Best model $x^*$
\end{algorithmic}
\end{algorithm}


\begin{figure}[t]
  \centering
  {\includegraphics[width=\linewidth]{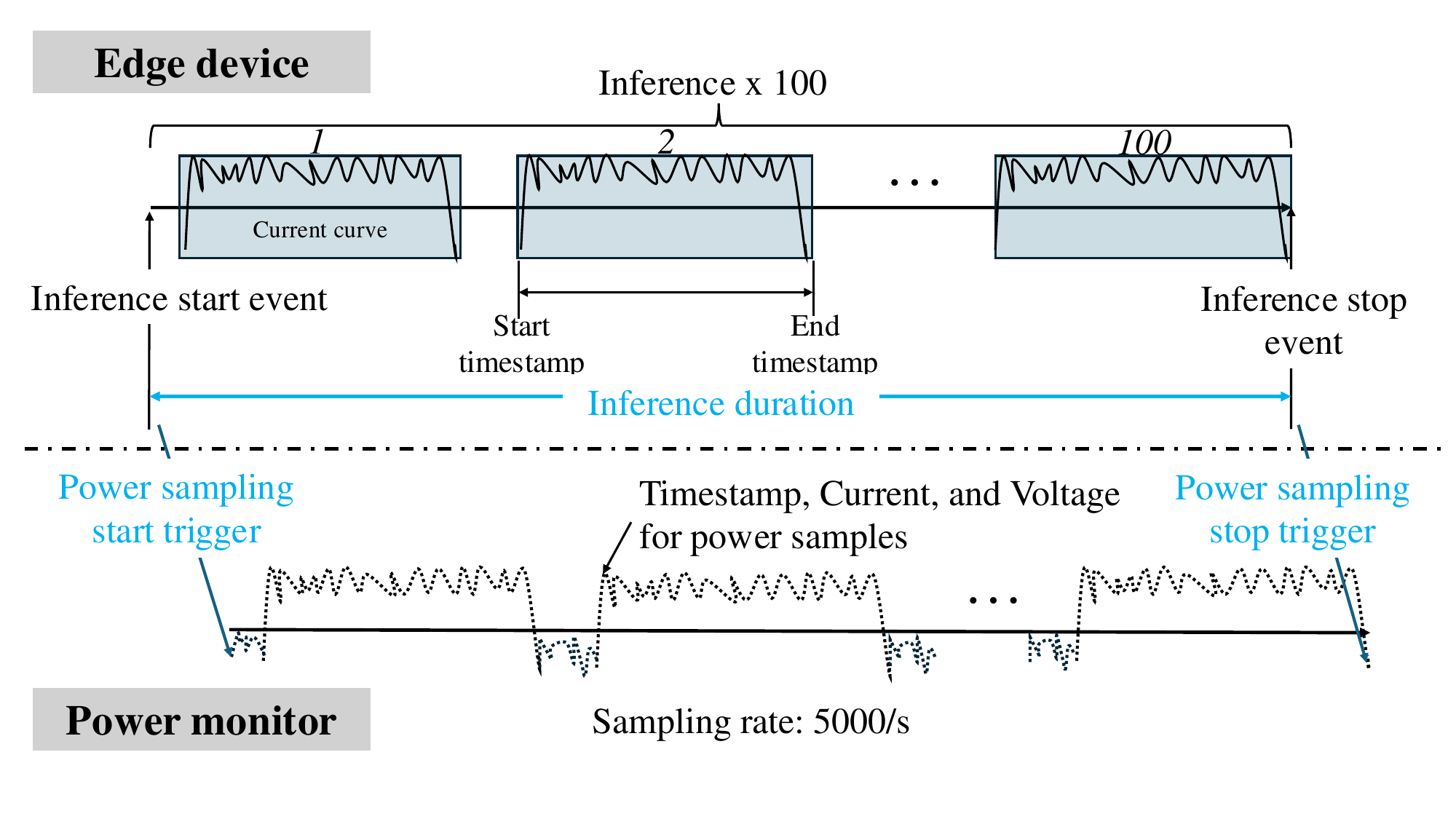}}
  \caption{Timing sync between edge device and external power monitor}
  \Description{system pipeline}
  \label{fig:time sync}
\end{figure}

\subsection{On-Device Measurement}
\label{ssc:measure}



As presented in \S \ref{ssc:Pareto}, iterative updates of the Pareto front require on-device energy measurements and model training.
\textsc{GreenAuto}, integrated with an external power monitor, automates this process, significantly improving efficiency compared to time-consuming manual measurements.
For different hardware, the energy consumption of the same model execution varies significantly. This variability arises because energy consumption is influenced by a multitude of hardware characteristics, including memory size, clock frequency, bandwidth, thermal efficiency, power management features, and more. Each of these factors contributes uniquely to the overall energy profile of a device, creating distinct energy footprints even for identical tasks.
In \textsc{GreenAuto}, the power consumption data sampling process is controlled either by a counter or timer. Since the model inferences are executed on the edge device and the power data is captured by the power monitor, proper synchronization between the two devices is critical yet non-trivial. As illustrated in Fig. \ref{fig:time sync}, the inference runs continuously on the edge device. For each inference, the edge device records the start and end timestamps, $T_{s}$ and $T_{e}$, as references. On the power monitor, power samples are exported, including three items: ${timestamp (T_{m}), current, voltage}$ at a fixed sampling rate. Directly synchronizing $T_{m}$ with $T_{s}$ and $T_{e}$ is challenging, as even when both devices use the same timing source, timing drift between the two can occur. This drift increases over time as sampling continues. Moreover, since the inference latency varies for each model, using a fixed timer to stop the sampling process is not feasible.
To address this, \textsc{GreenAuto} establishes two trigger events: the inference start event and the inference stop event, which control the sampling process. When inference starts, the start event triggers the power data sampling. When inference ends, the stop event terminates the sampling. This ensures that power samples always cover the entire inference duration. By using the start timestamp to identify the first sample and considering the total inference duration, we select all relevant power samples. The average current and voltage during inference are calculated from these selected samples, and the average power and total energy consumption are then determined based on the average current, voltage, and inference latency.
For accuracy, the model is trained and validated using application datasets, such as CIFAR-10 \cite{krizhevsky2009learning}, on the host GPU server.

\subsection{Reconfigurability}
\label{ssc:reconfigurability}
\textsc{GreenAuto} is highly flexible, enabling researchers to easily adapt the platform to their specific research objectives. In this paper, we demonstrate that by reconfiguring model structural parameters, we can effectively explore energy-efficient models. Similarly, researchers can define different parameter ranges and search spaces to focus on other objectives, such as maximizing accuracy or minimizing latency. 
For the model search process, parameters like the initial number of selected models ($m$) and the number of models chosen per iteration ($k$) can be fine-tuned to assess search efficiency. Additionally, search weights ($ws_i$) can be modified to control the optimal gradient direction, biasing the search toward specific performance goals. Model selection weights ($wd_i$) can also be applied to identify the best model for various scenarios, tailoring the selection from the Pareto front to meet diverse requirements.
\textsc{GreenAuto} is further customizable by allowing researchers to configure the hardware on which the models are deployed, acknowledging that performance and energy consumption vary across different platforms.


\section{System Implementation}

\begin{figure*}[t]
  \centering
  \begin{subfigure}{0.24\textwidth}
    \centering
    \includegraphics[width=\linewidth]{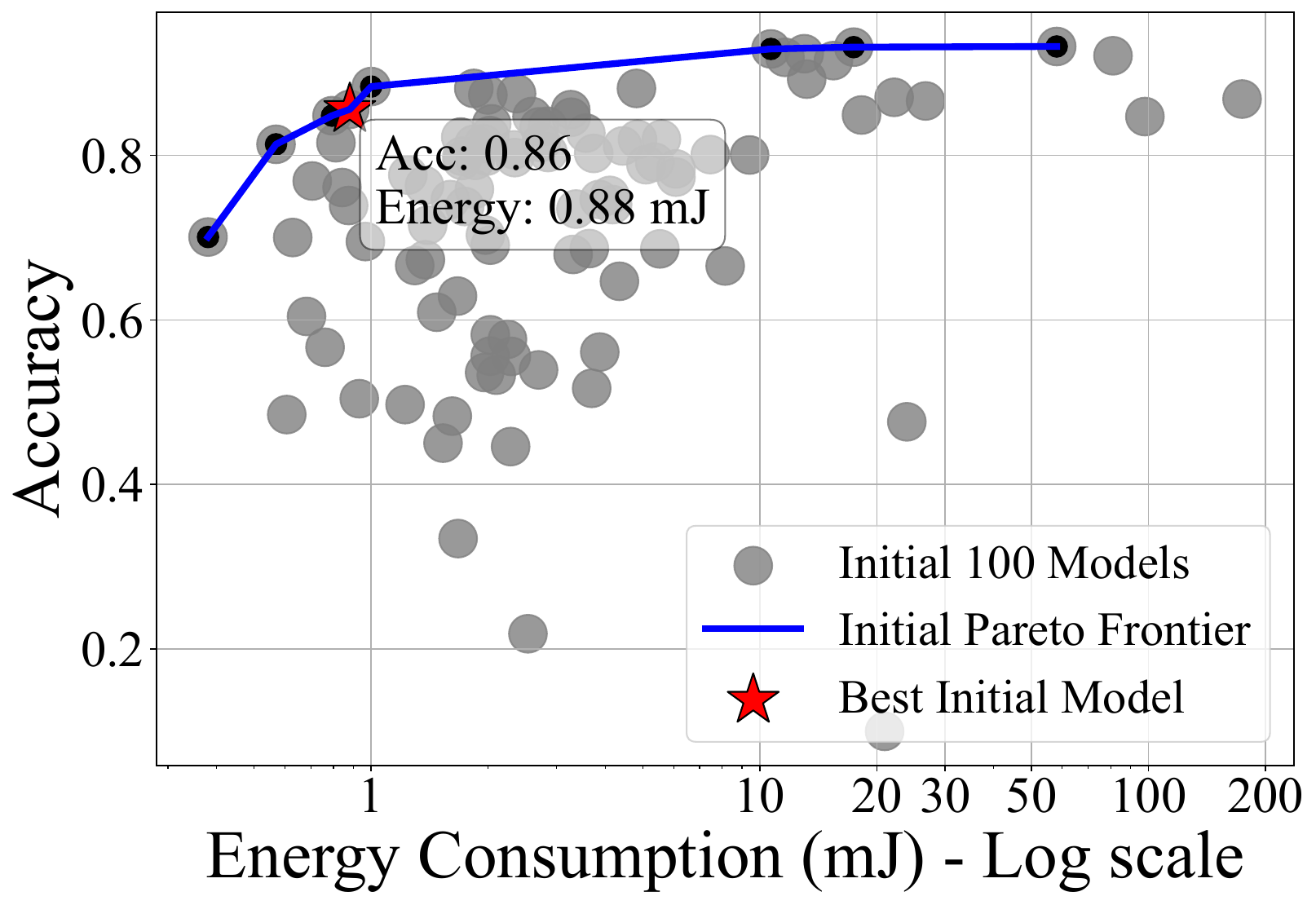}
    \caption{}
    \label{fig:iteration0}
  \end{subfigure}
  \begin{subfigure}{0.24\textwidth}
    \centering
    \includegraphics[width=\linewidth]{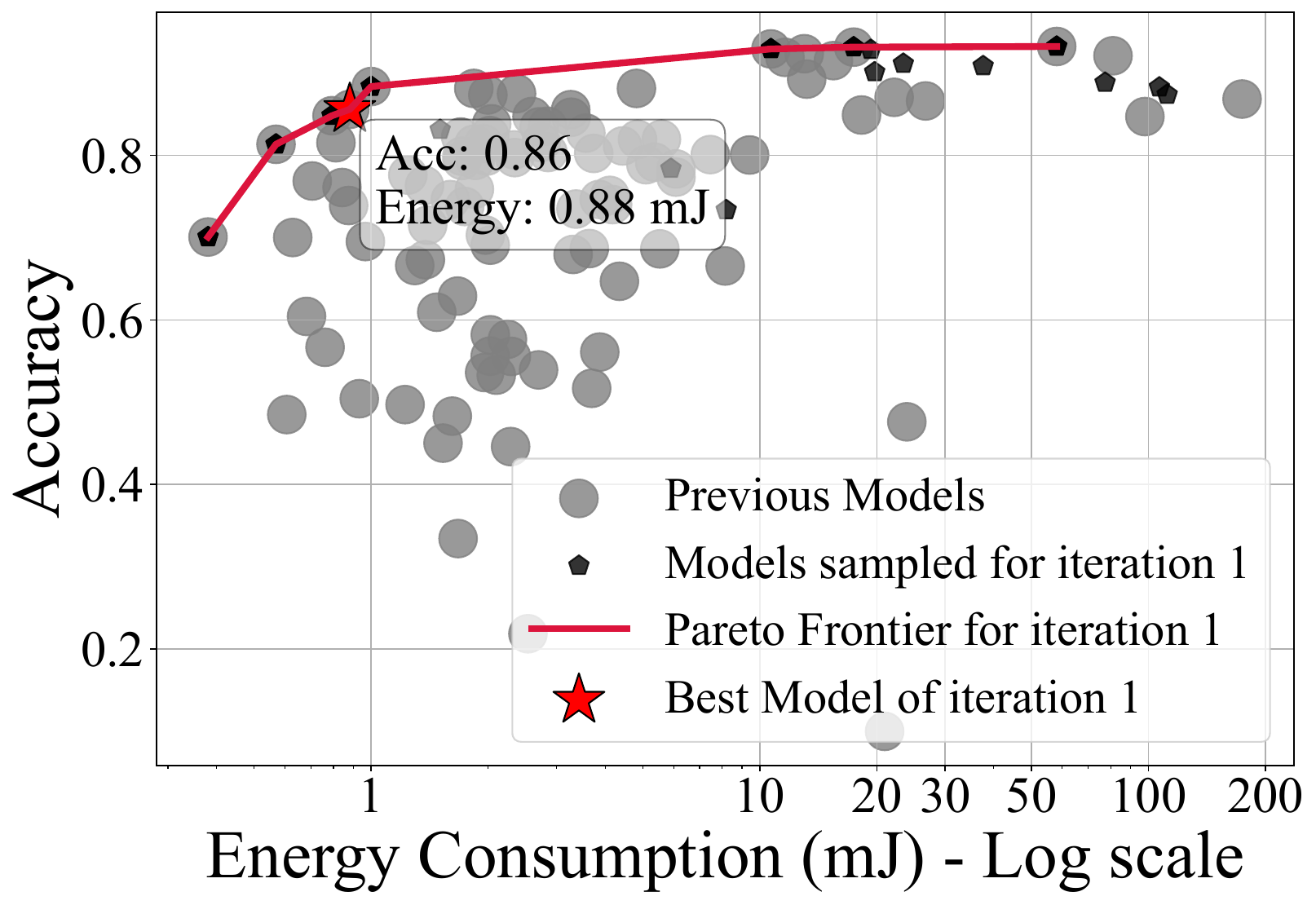}
    \caption{}
    \label{fig:iteration1}
  \end{subfigure}
  \begin{subfigure}{0.24\textwidth}
    \centering
    \includegraphics[width=\linewidth]{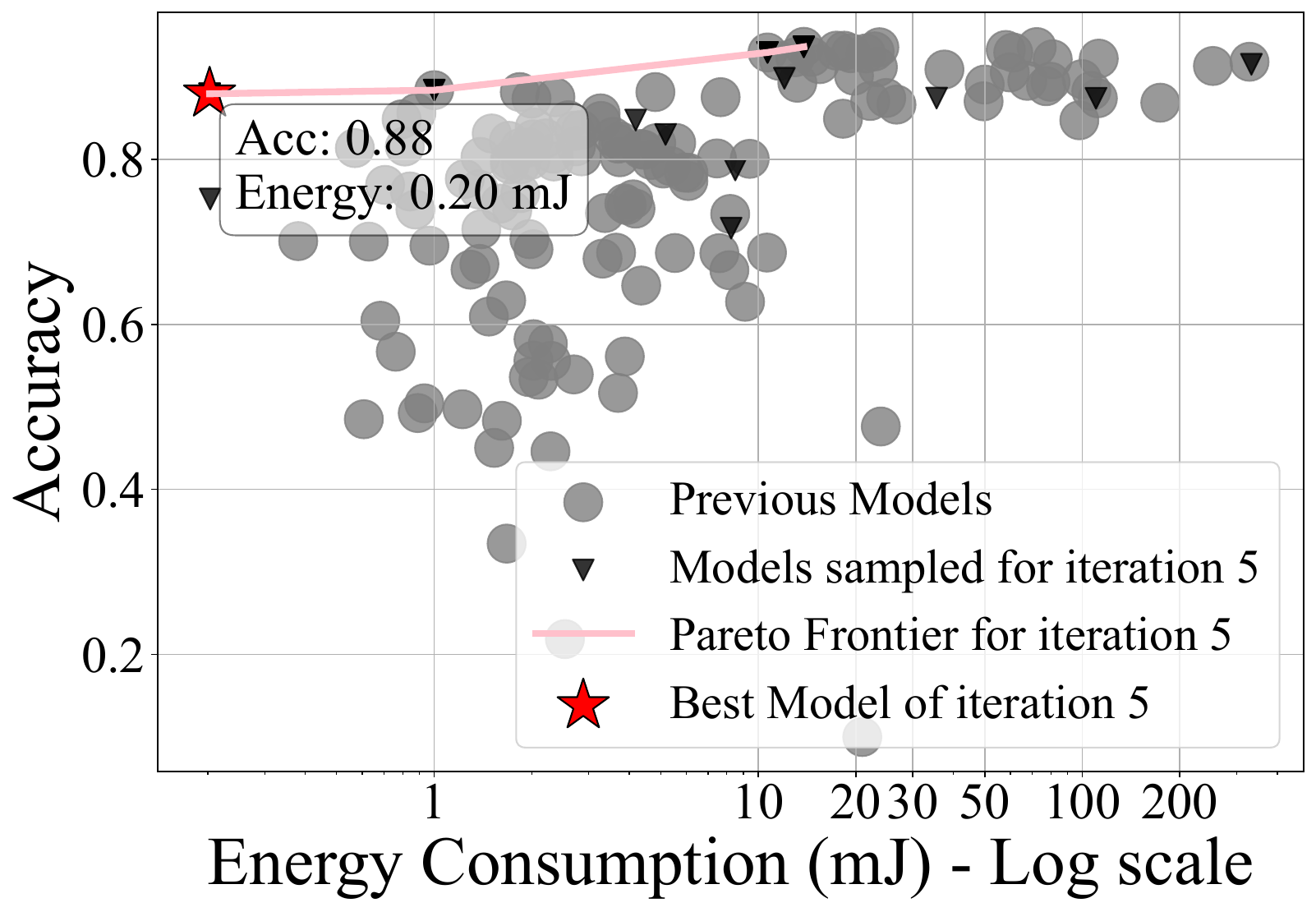}
    \caption{}
    \label{fig:iteration5}
  \end{subfigure}
  \begin{subfigure}{0.24\textwidth}
    \centering
    \includegraphics[width=\linewidth]{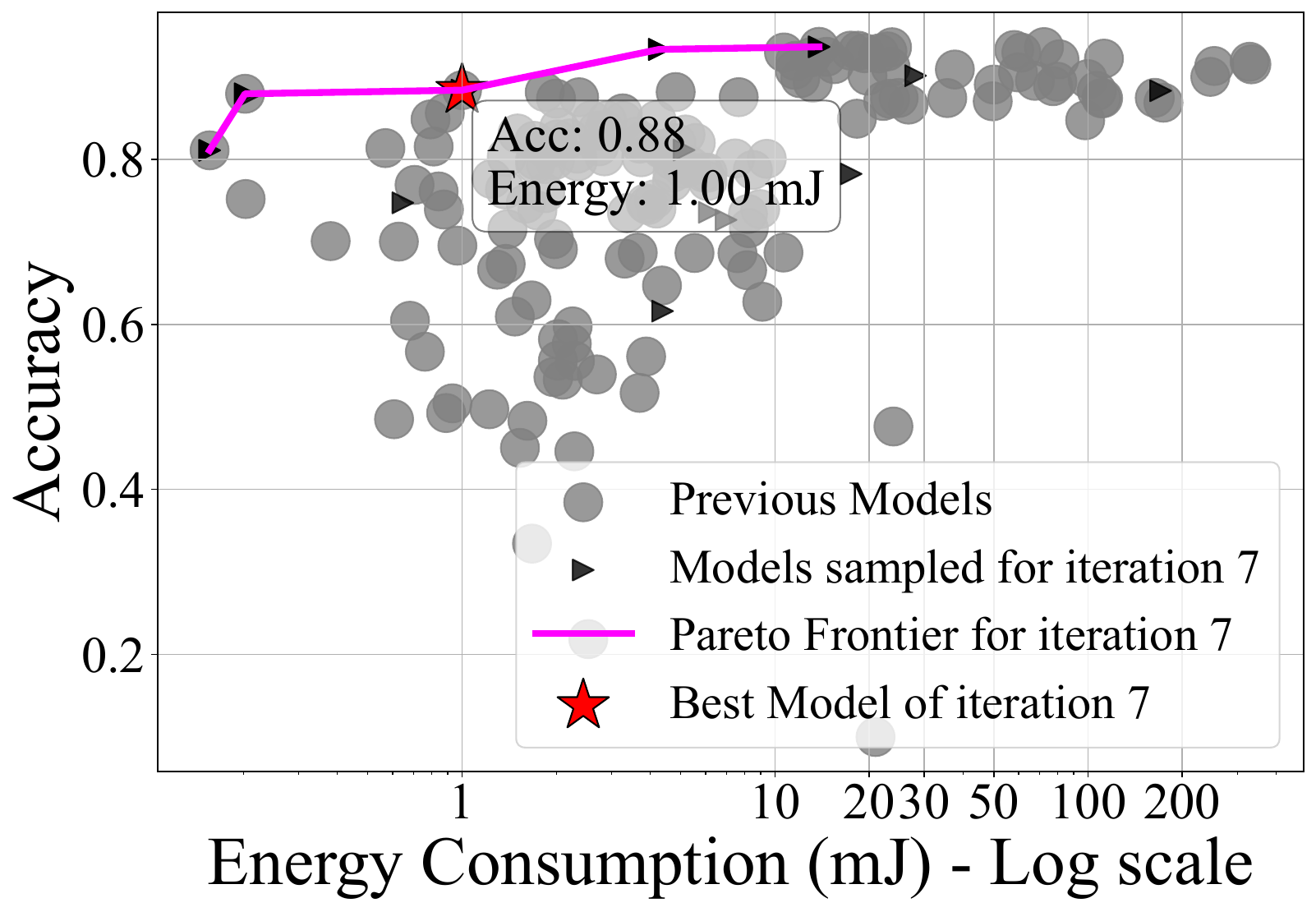}
    \caption{}
    \label{fig:iteration7}
  \end{subfigure}

  \caption{Pareto front update and the best model selection in each iteration.}
  \label{fig:iterations}
\end{figure*}

\textsc{GreenAuto} comprises four main hardware components: a host server, an edge device, a power monitor, and a Wi-Fi router, working together to form a fully automated system.

\textbf{GPU host server.} 
A GPU server is utilized for model generation, energy prediction, accuracy estimation, model search, model training, and model conversion. We use a single GTX 4090 to train the models and obtain their actual accuracy. For each model, the inference profiling logs and power consumption data collected from the edge device and power monitor are stored on the host server.

\textbf{Edge device.}
We run DNN model inference on an Android smartphone with an ARM Cortex A77 CPU and Mali A78 GPU, connected to the host server via Wi-Fi for automated model deployment. To ensure consistent power measurements, the device is powered directly through external power monitor, with its built-in battery removed. We standardize the environment by minimizing display brightness and disabling background services. Power consumption is measured using these precautions to ensure reliability across devices. For inference and latency profiling, we utilize the TensorFlow benchmark tool \cite{tflitebenchmarktool}, transmitting models and logs via the ADB bridge.


\textbf{Power monitor.}
The Monsoon power monitor \cite{Monsoon} is used as the external power monitor to collect power data. 
Given that models running on edge devices are typically small, with inference latencies in the microsecond range, the Monsoon's high sampling rate of 5000 Hz enables the capture of fine-grained power consumption during model execution. This offers significantly more precise measurements compared to software-based profilers.
Modern mobile devices often feature built-in batteries protected by a Battery Management System (BMS), which can complicate powering the device externally while ensuring it boots up properly. To address this, the most effective approach is to isolate the BMS chip from the device's battery without physically disassembling it. The BMS chip can then act as a bridge, connecting the device to the external power monitor, thereby maintaining safe operation and enabling precise power measurements without compromising the device's functionality.

\textbf{Wi-Fi router.}
A Wi-Fi router is used to connect the host server and the edge device. To ensure a stable connection, we use a dedicated router exclusively for communication between the host server and the edge device.

\section{Evaluation}
In this section, we evaluate \textsc{GreenAuto} through demonstrating how the search process evolves and how it selects the best model after each iteration. 
We set the initial model number $m = 100$ to generate the initial Pareto front. In each iteration, $k = 10$ new models are selected for on-device energy measurement. The model search weights, $ws_i$, are set to 1 for accuracy ($i=a$) and 3 for energy consumption ($i=e$), steering the search process toward energy-efficient models.
The stopping criterion for the search is defined as follows: if any model on the Pareto front achieves an accuracy greater than 0.9 on CIFAR-10 and energy consumption below 7 mJ, the search is terminated. This threshold is based on the performance of MobileNet-V2, which, when trained and measured on our platform, achieves an accuracy of 0.88 and energy consumption of 7.53 mJ.

\textbf{Model search process.} The model search and selection results are illustrated in Fig. \ref{fig:iterations}. The gray points represent the distribution of the initial 100 models, with energy consumption on the x-axis and accuracy on the y-axis. The blue line indicates the Pareto front constructed from these models, and the red star marks the best model selected based on Algorithm \ref{algorithm:BestModelSelection}. After the first iteration, the best model achieves an accuracy of 0.86 and an energy consumption of 0.88 mJ. 
In the next iteration, as shown in Fig. \ref{fig:iteration1}, 10 new models are selected and measured. The real measurements are used to update the Pareto front, represented by the red line. 
In the first iteration, neither the Pareto front nor the best model changes.
This process is repeated iteratively, selecting models to update the Pareto front and identifying the best models from the updated front. Ultimately, the best model we discovered achieves an accuracy of 0.88 and an energy consumption of only 0.2 mJ, as illustrated in Fig. \ref{fig:iteration7}.

\textbf{Best model selection.} Based on the same model search outputs, we compare Gradient Descent (GD), outlined in Algorithm \ref{algorithm:BestModelSelection}, with Weighted Sum (WS) to identify the best models. Comparison results are shown in Table.\ref{tab:model_selection}. 
For the Gradient Descent method, we tested two configurations. In the first, we set the model selection weights $wd_i$ to 10 for accuracy ($i=a$) and 1 for energy consumption ($i=e$), prioritizing accuracy. In the second, we set $wd_i$ to 1 for accuracy and 10 for energy consumption, prioritizing energy efficiency.
In the Weighted Sum method, each objective is multiplied by a corresponding weight, and their sum is used to evaluate the models. The model with the highest weighted sum is selected as the best. We used balanced weights in the WS method to give equal importance to both accuracy and energy efficiency.
The results demonstrate that the best model varies depending on the selection method and the weights applied. For instance, using GD, \textsc{GreenAuto}  identifies a model with an accuracy of 0.88 and significantly lower energy consumption of 0.2 mJ. While, with WS, \textsc{GreenAuto} selects a model with a higher accuracy of 0.93 and an acceptable energy consumption of 4.27 mJ through WS. Both models outperform MobileNet-V2 on the same edge device.

\begin{table}[t]
\centering
\caption{NAS search results on CIFAR-10.} 
\footnotesize
\resizebox{0.48\textwidth}{!}{
\begin{tabular}{c|c|c|c|c|c}
\toprule
Method & $wd_a$ & $wd_e$ & \makecell{Acc.$	\uparrow$} & \makecell{Inf. energy$	\downarrow$  (mJ)} & \makecell{Carbon$	\downarrow$ (kgCO$_2$)} \\
\hline
\multirow{2}{*}{\textsc{GreenAuto-GD}} & 10 & 1 & 0.88 & 0.2 & \multirow{3}{*}{\textbf{0.013}/model} \\  
\cline{2-5}
 & 1 & 10 & 0.81 & \textbf{0.16} &  \\
\cline{1-5}
GreenAuto-WS & 1 & 1 & \textbf{0.93} & 4.27 &  \\
\hline \hline
\multicolumn{3}{c|}{MobileNet-V2}  & 0.88 & 7.53 & - \\
\hline
\multicolumn{3}{c|}{NASNet-A}  & 0.92 & - & 0.231/model \\  
\bottomrule
\end{tabular}}
\label{tab:model_selection}
\end{table}

\textbf{Sustainability.} We compare carbon footprint of the search process and validation performance of the best models identified by \textsc{GreenAuto} against NASNet-A \cite{real2019regularized} in Table.\ref{tab:model_selection}. 
In this experiment, \textsc{GreenAuto} trains 170 models on a RTX 4090 GPU before the search process terminated. The process took nearly one week, showcasing the significant efficiency improvements. The best model achieves an accuracy of 0.93, with a carbon footprint of 0.013 kgCO$_2$ per model during search process. In contrast, NASNet-A requires $450$ Nvidia K40 GPUs running continuously for seven days to train 20,000 models, with the best model achieving a slightly lower accuracy of 0.92. However, its search process produces a significantly higher carbon footprint, 0.231 kgCO$_2$ per model. 
\textsc{GreenAuto} produces only $0.0478\%$ of the carbon emissions compared to NASNet-A, while achieving a slightly higher accuracy, demonstrating a substantial improvement in sustainability. 
The search process does require power, as GPU computations, data handling, and model evaluations consume energy. However, \textsc{GreenAuto} accounts for this energy consumption in its carbon footprint calculation, ensuring that the reported emissions accurately reflect the entire search process. By integrating a Pareto-based search strategy with energy predictors and an accuracy proxy, \textsc{GreenAuto} provides a comprehensive view of the search space, enabling a more efficient and targeted optimization process.
While the search process inherently involves energy consumption, \textsc{GreenAuto} is specifically designed to minimize its total environmental footprint. The sustainability gains achieved during the model deployment phase far outweigh the energy costs incurred during the search process. Ultimately, \textsc{GreenAuto} is not only effective in identifying accurate and energy-efficient models but also promotes environmental sustainability throughout the model search and deployment lifecycle.

\section{Discussion And Future Work}
\textsc{GreenAuto} offers an efficient end-to-end platform for researchers to explore sustainable AI models. Its flexible configurations make it easily adaptable to various research areas or development objectives. However, it has certain limitations.

\textbf{More comprehensive search space exploration.} 
Building on our previous work, we expanded the model configuration parameters to create a broader, energy efficiency-driven search space. However, the current search space remains somewhat constrained by its monotonous model type with a fixed skeleton. As state-of-the-art models, such as transformer-based models \cite{amatriain2023transformer} and LLaMA models \cite{touvron2023llama}, become increasingly prevalent on edge devices, assessing their energy efficiency becomes a critical task. Understanding and exploring their energy consumption and sustainability will be key for future advancements in this field.

\textbf{Integration of hardware heterogeneity.}
As edge devices continue to evolve, they incorporate diverse hardware architectures, including CPUs, GPUs, TPUs, and AI accelerators. Integrating support for hardware heterogeneity into our platform is essential to ensure it remains adaptable and efficient. Tailoring model deployment to harness the unique capabilities of each hardware type will allow for the exploration of co-design strategies, optimizing energy-efficient hardware-software combinations for edge devices.

\section{Acknowledgment}
This work was supported by funds from Toyota Motor North America and sponsored by the Army Research Laboratory under Cooperative Agreement Number W911NF-23-2-0224. The views and conclusions contained in this document are those of the authors and should not be interpreted as representing the official policies, either expressed or implied, of the Army Research Laboratory or the U.S. Government. The U.S. Government is authorized to reproduce and distribute reprints for Government purposes notwithstanding any copyright notation herein. We thank the reviewers for their constructive criticism and valuable advice, which have improved the quality of this final paper.

\bibliographystyle{unsrt}
\bibliography{references}
\end{document}